\documentclass[fleqn,10pt]{wlscirep}
\usepackage{url}
\usepackage[colorlinks=true]{hyperref}
\usepackage{comment}
\usepackage{float}
\usepackage{hyphenat}
\usepackage{adjustbox}
\usepackage{longtable}

\usepackage{tikz}

\PassOptionsToPackage{square,numbers}{natbib}

\usepackage[utf8]{inputenc} 
\usepackage[T1]{fontenc}    

\usepackage{xurl}
\usepackage{url}            
\usepackage{booktabs}       
\usepackage{amsfonts}       
\usepackage{nicefrac}       
\usepackage{microtype}      
\usepackage{authblk}
\usepackage{multirow}
\usepackage{graphicx}
\usepackage{pifont}
\usepackage{array}

\newcommand{\cmark}{\ding{51}}%
\newcommand{\xmark}{\ding{55}}%

\usepackage{wrapfig,lipsum,booktabs}
\usepackage{tabularx, multirow, makecell, booktabs, caption}
\usepackage{ragged2e}
\title{SeasFire as a Multivariate Earth System Datacube for Wildfire Dynamics}

\author[1]{Ilektra Karasante} 
\author[2]{Lazaro Alonso}
\author[1,3]{Ioannis Prapas}
\author[1,4]{Akanksha Ahuja}
\author[2,5,6]{Nuno Carvalhais}
\author[1,7]{Ioannis Papoutsis}

\affil[1]{National Observatory of Athens, Institute for Astronomy, Astrophysics, Space Applications and Remote Sensing, Penteli, 15236, Greece}
\affil[2]{Max Planck Institute for Biogeochemistry, Department of biogeochemical Integration, Jena, 07745, Germany}
\affil[3]{Universitat de València,Image Processing Laboratory, València, 46022, Spain} 
\affil[4]{University of Cambridge, Department of Engineering, Cambridge, CB2 1PZ, United Kingdom} 
\affil[5]{Universidade Nova de Lisboa, Departamento de Ciências e Engenharia do Ambiente, Faculdade de Ciências e Tecnologia, Caparica, 2829-516, Portugal} 
\affil[6]{ELLIS Unit Jena, Jena, Germany} 
\affil[7]{National Technical University of Athens, School of Rural, Surveying and Geoinformatics Engineering, Zografou, 15773, Greece} 

\affil[*]{corresponding author: Ilektra Karasante (corresponding. ile.karasante@noa.gr)}

\begin{abstract}

The global occurrence, scale, and frequency of wildfires pose significant threats to ecosystem services and human livelihoods. To effectively quantify and attribute the antecedent conditions for wildfires, a thorough understanding of Earth system dynamics is imperative. In response, we introduce the SeasFire datacube, a meticulously curated spatiotemporal dataset tailored for global sub-seasonal to seasonal wildfire modeling via Earth observation. The SeasFire datacube comprises of 59 variables encompassing climate, vegetation, oceanic indices, and human factors, has an 8-day temporal resolution and a spatial resolution of 0.25  $^{\circ}$, and spans from 2001 to 2021. We showcase the versatility of SeasFire for exploring the variability and seasonality of wildfire drivers, modeling causal links between ocean-climate teleconnections and wildfires, and predicting sub-seasonal wildfire patterns across multiple timescales with a Deep Learning model. We publicly release the SeasFire datacube and appeal to Earth system scientists and Machine Learning practitioners to use it for an improved understanding and anticipation of wildfires.  

\end{abstract}

\begin{document}

\flushbottom
\maketitle

\thispagestyle{empty}

\section*{Background \& Summary}

Wildfires, as integral components of terrestrial ecosystems, play a significant role in shaping ecological development through disturbance and regeneration \cite{mclauchlan_2020_fire}. However, the increasing influence of both climate change and human activities has disrupted the natural fire cycle and modified ecosystems \cite{unitednationsenvironmentprogramme_2022_number, harper_2018_prescribed}. Considering the expected significant changes in climate over the next century, it is imperative to reevaluate wildfire adaptation and mitigation strategies \cite{jolly_2015}. In terms of impact, wildfires exert a significant ecological influence by enhancing nutrient cycling, initiating ecological succession, creating diverse habitats, and supporting fire-adapted species \cite{steel_2019, Schoennagel_2017}. However, when wildfires exceed expected intensity or frequency, they can have devastating impacts on ecosystem services, infrastructure, communities, and public health \cite{cas_2018, liu_2015, holm_2020, Beranek_2023, xu_2023}. 

To effectively mitigate wildfires, a robust characterization of the complex dynamics spanning atmospheric, oceanic, and terrestrial processes is imperative. Recognizing the urgency inherent in fire prediction and management, we synthesize a novel dataset, named SeasFire, which paves the way for the development of data-driven methods for forecasting wildfire patterns and impacts on a sub-seasonal scale, incorporating teleconnections. The dataset comprises of 59 global variables covering climatic, meteorological, environmental, and anthropogenic wildfire drivers, along with historical burnt areas and carbon emissions, spanning over two decades, from 2001 to 2021. These variables are distributed within a 0.25 $^{\circ}$ resolution striking a balance between offering detailed spatial information, ensuring computational efficiency, and facilitating compatibility with existing data sources. The SeasFire datacube has an 8-day temporal resolution, effectively capturing both short-term fluctuations and maintaining a sufficient number of data points across a 21-year duration. This 8-day interval is particularly well-suited for capturing environmental phenomena characterized by seasonal or sub-seasonal cycles.

There have been significant efforts to consolidate wildfire datasets, as illustrated in Table \ref{tab:table1}, which delineates regions, temporal extents, and characteristics for each dataset. While these datasets provide valuable insights into specific aspects of wildfires, a notable gap exists in the availability of multivariate Earth Data Cubes dedicated to wildfires. Additionally, these datasets have predominantly focused on specific regions within the United States of America and Europe, leading to two consequential limitations. Firstly, the absence of globally scaled variables impedes the ability to model Earth system processes and dynamics affecting wildfires on a global scale. This limitation hinders a comprehensive understanding of the interconnectedness of these processes across different regions. Secondly, the potential positive impacts of anticipating seasonal wildfires are constrained in resource-efficient countries and regions.
SeasFire's global coverage not only enables a more comprehensive understanding of global wildfire dynamics but also contributes to the development of more robust and location-agnostic models. Such models have the potential to generalize well across diverse geographical areas, transcending the limitations posed by regional-focused datasets.

Consolidating a comprehensive, global, time-series dataset for wildfire modeling using Earth Observation variables poses a significant challenge, as navigating through the ever-expanding data landscape presents formidable hurdles.
Indeed, the domain of Earth Observation is undergoing exponential growth, evidenced by the availability of over 400 publicly accessible datasets \cite{EarthNets}. These datasets, hosted on various geospatial platforms, such as the Climate Data Store (CDS), Google Earth Engine, and the Copernicus Open Access Hub, collectively amass hundreds of terabytes of data \cite{Giuliani2019}. They offer diverse spatiotemporal scales and modalities, forming a rich repository of information.
This wealth of data is distributed across multiple services and repositories. Adding to the complexity, data is collected from a diverse array of sensors such as Landsat, MODIS, and Sentinel, each possessing unique characteristics and capabilities.
To harness the potential of such extensive datasets for deriving valuable insights and developing accurate wildfire models, the scientific community must adeptly handle demanding tasks of data selection, access, and harmonization. These challenges can be effectively addressed through the utilization of datacubes \cite{Baumann2018, Mahecha2020, MonteroLoaiza2023}. Datacubes offer a streamlined approach to managing and analyzing large, multi-dimensional Earth Observation datasets, providing a unified framework for efficient data exploration and model development.

SeasFire cube design is a scientific asset based on a novel paradigm for data-driven wildfire research using analysis-ready data. The cube specifications and granularity allow the development of Earth system deep learning models for wildfire science and beyond, that capture the long spatiotemporal interactions of Earth system variables. Distinguished by its incorporation of ocean climate indices, SeasFire can be used to probe the Earth's spatiotemporal interactions, such as memory effects and teleconnections to capture the dynamic and non-linear interactions of the Earth system components, particularly in the context of seasonal wildfire forecasting. This innovative dataset empowers researchers in the Earth system sciences, facilitating rigorous analytical and predictive modeling on both regional and global scales. In addition, as a cloud-friendly dataset, it also tackles computational challenges and removes storage constraints for seamless data analysis.Beyond wildfires, SeasFire enables the study of environmental phenomena such as vegetation dynamics and drought monitoring. Researchers can use the SeasFire datacube to explore the spatiotemporal distribution of wildfire carbon emissions, track changes, and identify sources and sinks, acknowledging variations based on fuel types.

The dataset has already been used for modeling global wildfire patterns with deep learning models \cite{Prapas_2022, Prapas_2023}. Prapas et al. (2022)\cite{Prapas_2022}, use semantic segmentation on the SeasFire cube for burned area pattern forecasting. An extension of this work is TeleVit\cite{Prapas_2023}, a transformer model that captures teleconnection information to improve performance at larger forecasting horizons. TeleViT combines local views at higher resolution ($0.25^{\circ}$), global views at lower resolution ($1^{\circ}$), and time-series of ocean-climate indices, a setting that is allowed by the versatile structure of the SeasFire datacube. Moreover, the Pi-school organization has employed the SeasFire datacube to comprehend sub-seasonal to seasonal forecasts of global burned areas, harnessing explainable artificial intelligence (AI) techniques integrated with deep learning models. Their work can be found in the GitHub repository (\url{https://github.com/PiSchool/noa-xai-for-wildfire-forecasting}). 

\begin{table} [ht!]
\begin{longtable}{|p{1.75cm}|p{10.5cm}|p{1cm}|p{1.2cm}|p{1cm}|}
\hline
\textbf{Name} & \textbf{Description} & \textbf{Fire Drivers} & \textbf{Region} & \textbf{Years} \\
\hline
\endhead
MODIS \cite{nasaFIRMSFrequently}  & The 1km MODIS active fire product detects thermal anomalies using a contextual algorithm and provides location, brightness temperature and spatio-temporal attributes. & \xmark & GLB & 2000 - Present \\
\hline
VIIRS \cite{nasaFIRMSFrequently} & The 375m VIIRS active fire product detects fires with improved resolution and nighttime performance compared to MODIS and adds fire radiative power. & \xmark & GLB & 2012 - Present \\
\hline
FireCCI \cite{esa_fire_cci} & The FireCCI burned area product provides a sum of the burned area, standard error, fraction of burnable area, fraction of observed area, number of patches, and a sum of the burned area for each land cover category. & \xmark & GLB & 2001 - Present\\
\hline
Fire Atlas \cite{essd-11-529-2019} & The Global Fire Atlas tracks the dynamics of individual fires to determine the location and timing of ignitions, duration and size of fires, daily expansion of fires, along with line length, speed, and direction of fire spread. & \xmark & GLB & 2003 - 2016 \\
\hline
NIFC  \cite{arcgisHistoricPerimeters} & The geospatial dataset includes data on the boundaries of wildfires and offers historical information in the United States, including fire boundaries, size, and other attributes using GeoMAC. & \xmark & US & 2000 - 2018 \\
\hline
GeoMAC \cite{arcgisNationalInteragency} & Geospatial Multi-Agency Coordination Group (GeoMAC) is an internet-based mapping tool that was initially created for fire managers to obtain real-time fire perimeter data for active wildfires facilitating immediate monitoring and management. & \xmark & US & 2000 - 2019 \\
\hline
CNFDB \cite{nrcanCanadianWildland} & Canadian National Fire DataBase serves as a comprehensive repository of wildfire information in Canada, including fire location, size, cause, and suppression details. & \xmark & CAN & 1980 - Present \\
\hline
EFFIS \cite{europaEFFISWelcome} & Acts as a central hub for wildfire information and datasets in Europe, offering data on fire events, burned areas, and fire danger indices. & \xmark & EU & 2015 - Present \\
\hline
Bushfire \cite{gaProductCatalogue} & Offers vital information on wildfires in Australia, including fire history, severity, and vegetation data. & \cmark & AUS & 1900 - Present \\
\hline
Wildfires Australia \cite{arcgisArcGISDashboards} & Amount, size and anomalies of surface temperature of wildfires on an interactive dashboard & \cmark & AUS & 2011 - 2020 \\
\hline
Kaggle \cite{short2017spatial} & Provides information on different fire sizes, frequencies, and causes, offering a comprehensive resource for studying 1.88 million wildfires in the US. & \xmark & US & 1992 - 2015 \\
\hline
Sentimental Wildfire \cite{lever2022sentimental} & Integrates geophysical satellite data from the Global Fire Atlas with Twitter's social data and applies sentiment analysis to social media for more accurate predictions of wildfire characteristics. & \cmark & US, AUS & 2016 \\
\hline
Incidents \cite{st2023all} & Captures various hazards, with fire-related incidents constituting the majority, highlighting the significance of fire-related events. & \xmark & US & 1999 - 2014 \\
\hline
WildfireDB \cite{singla2021wildfiredb} & Encompasses 17 million data points, allowing in-depth understanding of fire spread dynamics in the continental USA over the past decade. & \xmark & US & 2012 - 2017 \\
\hline
FIRE-ML \cite{graff2021fireml} & Provides a daily wildfire forecasting dataset for the contiguous United States, including active fire detections, land cover, and more. & \cmark & US & 2012 - 2020 \\
\hline
\caption{\label{tab:table1}Summary of Wildfire Datasets (GLB (Global), US (United States), AUS (Australia), CAN (Canada), EU (Europe))}
\end{longtable}
\end{table}

\section*{Methods}

This section provides an overview of the dataset and all necessary curation steps undertaken for building an Earth system datacube for seasonal and sub-seasonal fire forecasting. Adhering to FAIR principles (Findable, Accessible, Interoperable, and Reusable) \cite{Wilkinson2019, Giuliani2019}, we designed and constructed the SeasFire datacube, ensuring that it offers a user-friendly experience for researchers and analysts, to discover, access, and leverage these variables for various applications. The datacube's architecture is designed to be flexible, accommodating the addition of extra variables, as well as including many variables that grant users the freedom to extend them as necessary over time.

\subsection*{Development of SeasFire datacube}
There are three foundational concepts essential to our work: datasets, datacubes (alternatively referred to as data cubes or simply cubes), and data arrays. A dataset functions as a container that accommodates multiple data variables, while a datacube represents a specialized form of dataset specifically designed for spatiotemporal data. Data arrays denote individual variables that may constitute a dataset or datacube. These concepts are frequently employed collectively, particularly in the context of managing complex, multi-dimensional data in scientific and geospatial applications.

Datacubes aim to tackle the challenges posed by Big Data through their cloud-optimized architectures \cite{Kopp2019, Sudmanns2021, ArizaPorras2017, Dhu2017, Lewis2017, Killough2018}. Datacube initiatives have become pivotal in the realm of Earth observation and data analysis, exemplified by the Committee on Earth Observation Satellites (CEOS) as a founding partner of the Open Data Cube initiative \cite{Killough2018} and a spectrum of diverse projects worldwide. Flagship projects in this domain include Digital Earth Australia \cite{Lewis2017, Dhu2017}, the Colombian Data Cube \cite{ArizaPorras2017}, the Swiss Data Cube \cite{Giuliani2017}, and several others. Furthermore, noteworthy initiatives include the Semantic Austrian EO Data Cube Infrastructure \cite{Sudmanns2021} and the Earth System Data Lab (ESDL) \cite{Mahecha2020}, purposefully designed to enhance the efficiency of analyzing analysis-ready data (ARD). CEOS defines ARD as “satellite data that have been processed to a minimum set of requirements and organized into a form that allows immediate analysis without additional user effort” \cite{Killough2018, Lewis2018}.  Additionally, customized regional initiatives like the Regional Earth System Data Lab \cite{EstupinanSuarez2021} further augment the datacube landscape. Furthermore, the existence of large-scale datacube initiatives on a global level, like the Earth System Data Cube \cite{Mahecha2020}, highlights the growing importance of this approach in the field of Earth observation.

Building a datacube can be achieved through various methods, with options ranging from dedicated software platforms such as the Open Data Cube (ODC) \cite{Killough2018} to Python libraries like xarray \cite{xarray} or its Julia equivalent YAXArray.jl, each fulfilling distinct roles. ODC primarily emphasizes data storage and retrieval, whereas xarray centers its focus on the manipulation of data within memory. Many research datacubes \cite{Mesogeos, kondylatos_wildfire_2022, Mahecha2020} tailored to specific research domains, have been developed by using xarray\cite{xarray} as a key component,  harnessing its capabilities for handling diverse data types, including climate data, environmental data, and scientific measurements. 

Consequently, by leveraging xarray we developed the SeasFire datacube, a harmonized spatiotemporal Earth system datacube capable of accommodating multiple datasets. Datacubes have the ability to conduct concurrent analyses, by effectively converting vast data volumes into easily accessible and valuable insights. To address the challenges of managing multidimensional arrays, we provide the cube in .Zarr specification \cite{zarr}. This is particularly well-suited for cloud-based environments, as it provides efficient chunk access mechanisms, facilitating seamless parallel processing of data. We exploit cloud data optimization to foster the exploration of the complex and interconnected dynamics of the multivariate Earth system, employing a standardized approach to generate customized datacubes instantly. This approach aims to help Earth system scientists and machine learning practitioners select the variables and dimensions essential for training their models. 

\subsection*{SeasFire datacube variables} 

 The SeasFire datacube offers a comprehensive collection of variables that capture key environmental factors associated with fire drivers as well as fire targets such as burned area, fire radiative power, and wildfire-induced carbon dioxide emissions.  In total, this datacube comprises 59 variables spanning various domains. These include climatic elements, featuring ocean-climate indices; atmospheric parameters, including temperature, pressure, and humidity-related variables; land-related aspects, like land cover and population density variables; and several masks as biomes and land-sea mask, all of which exert influence on fire behavior. These categories are organized as shown in table \ref{tab:table2}. Furthermore, the datacube includes distinct data arrays that capture various ways of summarizing a variable. For instance, it provides temperature data at a 2-meter height in maximum, minimum, and average values. Additionally, the datacube contains arrays of data that originate from breaking down information initially found in a combined variable. An example of this is the comprehensive land cover category that encompasses multiple classes. A process of aggregation and regridding, results in the creation of individual variables, each focused on representing a specific land cover class. Incorporating a comprehensive set of variables within the datacube, enriches the information available to machine learning models, enabling a more customized and understandable modeling of seasonal fire patterns.

\begin{figure}[htbp]
    \centering
    \includegraphics[width=0.5\columnwidth]{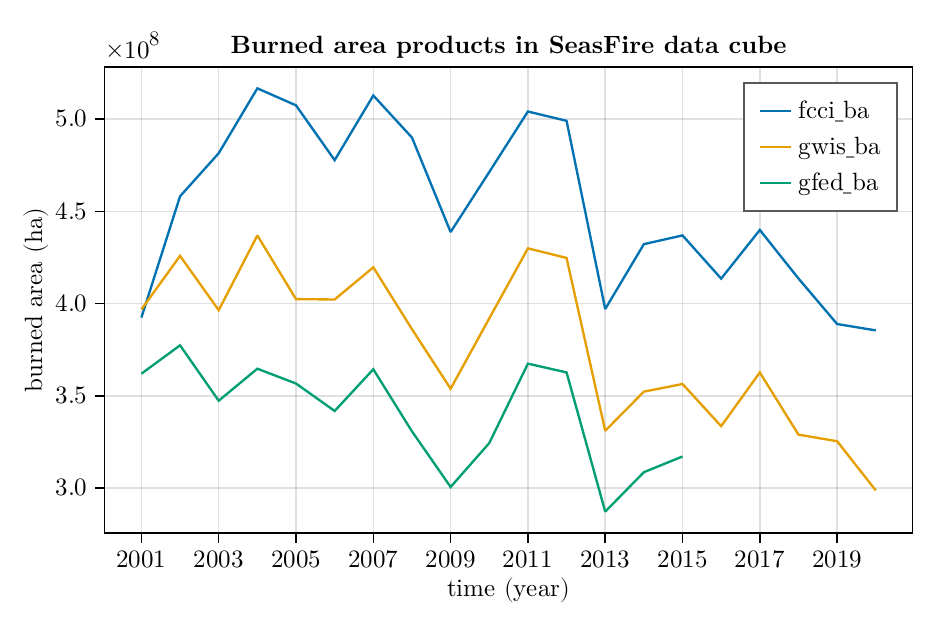}
    \caption{Global annual cumulative timeseries of the three burned areas products of the SeasFire datacube.}
    \label{fig:burned_areas}
\end{figure}

\paragraph*{Burnt areas variables} The datacube contains three different products related to burned areas: the GWIS\cite{gwis}, the GFED\cite{gfed}, and the FCCI\cite{fcci} dataset. Global burned area products still rely on moderate resolution sensors, with 250–500 m pixel sizes and 1–2 days revisit time\cite{LizundiaLoiola2022}. Despite Landsat and Copernicus Sentinel-2 satellites openly providing better resolutions (10-30m), the obstacle of extensive processing effort required to produce comprehensive, long-term global datasets remains \cite{LizundiaLoiola2022}. Therefore securing global, daily data on burnt areas has been a notable challenge. Since 2001, the MODIS Aqua and Terra satellites have been delivering consistent and quality-checked data, inspiring various organizations to compile global burnt areas datasets using a variety of data sources, methodologies, and spatiotemporal resolutions. A diverse range of burned area products has been developed using data from these satellite sensors. Some noteworthy examples include the MCD64A1 v6.0 dataset \cite{mcd64a1.006}, which provides spectral reflectance data at 500 meters resolution, along with MODIS 1-kilometer active fire data \cite{Giglio2009, Giglio2016, Giglio2018}, spanning from 2000 to the present day. Given the benefits of that product, numerous researchers and organizations have advanced their efforts to develop robust datasets for burned areas, each with its unique set of advantages and drawbacks. For instance, the GFED \cite{gfed} dataset provides daily data up to 2015 up to 500 meters resolution, while the FCCI \cite{fcci} (Fire Climate Change Initiative) dataset, developed by the European Space Agency (ESA), offers monthly data at 250 m resolution. Furthermore, the GWIS \cite{gwis} dataset has produced global shapefiles on a monthly scale along with the ignition dates, spanning from 2000 to 2020 at a 500-meter resolution.  Due to variations in data processing applied for each dataset, the burned area products and corresponding values in hectares are different (figure \ref{fig:burned_areas}). These differences are well-documented and explained in relevant literature \cite{Turco2019, Valencia2020}. By including those three datasets in the SeasFire datacube we allow researchers to select the one that aligns most effectively with their specific requirements. 

\subsection*{Assumptions \& Specifications}

The SeasFire datacube offers a spatiotemporal perspective that facilitates the examination of global fire patterns. Table \ref{tab:table2} summarizes the various dataset collections, including their spatiotemporal details. The data are organized in a three-dimensional grid, with each grid cell covering a spatial resolution of 0.25 $^{\circ}$ in latitude and longitude. A 0.25 $^{\circ}$ global grid aligns with various global datasets and standards, facilitating interoperability and compatibility with existing data sources (e.g. Climate data store, ESA CCI, NASA SEDAC) and at the same time has a fine resolution to capture variations in atmospheric and climatic data. Temporally, the datacube provides information aggregated over 8-day intervals, allowing for seasonal and sub-seasonal analysis and forecasting. The decision to use an 8-day resolution is deliberate, given the availability of the MODIS NDVI \cite{ndvi_ds} product in a 16-day resolution, making the 8-day interval well-suited for resampling. The 8-day granulation commences on January 1st of each year, resulting in an annual total of 46 datetimes. With a span of 21 years, the datacube allows for long-term trend analysis and exploration of interannual variability.

\begin{table}
\def\arraystretch{1.1}%
\centering
\begin{tabularx}{7.06in}{|p{1in}|p{1.2in}|p{2.5in}|p{1in}|p{0.5in}|}\hline
\textbf{Provider} & \multicolumn{1}{l|}{\textbf{Dataset}} & \textbf{Variables} & \textbf{Time Span/Res.} & \textbf{Spatial Res.}\\\hline

\RaggedRight{Climate Data Store}&
\RaggedRight{ERA5 hourly data on single levels from 1940 to present}&
\RaggedRight{Mean sea level pressure; Total precipitation; Relative humidity; Vapor Pressure Deficit; Sea Surface Temperature; Skin temperature; Wind speed at 10 meters; Temperature at 2 meters; Surface net solar radiation; Surface solar radiation downwards; Volumetric soil water levels 1,2,3 and 4}& 
\RaggedRight{2001-2021 / hourly}&
\RaggedRight{0.25 $^{\circ}$}
\\\hline

\RaggedRight{Climate Data Store}&
\RaggedRight{CEMS Global Fire Assimilation System Historical Data.}&
\RaggedRight{Drought Code; Fire Weather Index}& 
\RaggedRight{2001-2021 / hourly}&
\RaggedRight{0.25 $^{\circ}$}
\\\hline

\RaggedRight{Atmosphere Data Store}&
\RaggedRight{CAMS global biomass burning emissions based on fire radiative power (GFAS)}&
\RaggedRight{Carbon dioxide emissions from wildfires; Fire radiative power}& 
\RaggedRight{2003-2021 / daily averages}&
\RaggedRight{0.1 $^{\circ}$}
\\\hline

\RaggedRight{NOAA, National Oceanic, and Atmospheric Administration}&
\RaggedRight{Climate Indices: Monthly Atmospheric and Ocean Time Series}&
\RaggedRight{Western Pacific Index; Pacific North American Index; North Atlantic Oscillation; Southern Oscillation Index; Global Mean Land/Ocean Temperature; Pacific Decadal Oscillation; Eastern Asia/Western Russia; East Pacific/North Pacific Oscillation; Nino 3.4 Anomaly; Bivariate ENSO Timeseries; Arctic Oscillation}& 
\RaggedRight{2001-2021 / monthly}&
\RaggedRight{-}
\\\hline

\RaggedRight{ESA CCI}&
\RaggedRight{Land cover classification gridded maps from 1992 to present derived from satellite observations}&
\RaggedRight{No data; Agriculture; Forest; Grassland; Wetlands; Settlement; Shrubland; Sparse vegetation; bare areas, permanent snow and ice,  Water Bodies}& 
\RaggedRight{2001-2021 / monthly}&
\RaggedRight{300 m}
\\\hline

\RaggedRight{NASA LP DAAC at the USGS EROS Center}&
\RaggedRight{MOD11C3 v006 / MCD15A2H v006/ MOD13C1}&
\RaggedRight{Land Surface Temperature / Leaf Area Index / Normalized Difference Vegetation Index}& 
\RaggedRight{2001-2021 / 8 \& 16 days averages}&
\RaggedRight{0.05 $^{\circ}$ / 500 m / 0.05 $^{\circ}$ }
\\\hline

\RaggedRight{RESOLVE Biodiversity and Wildlife Solu\hyp{tions}}&
\RaggedRight{RESOLVE Ecoregions 2017}&
\RaggedRight{biomes}& 
\RaggedRight{static / -}&
\RaggedRight{-}
\\\hline

\RaggedRight{NASA SEDAC, Socioeconomic Data and Applications Center}&
\RaggedRight{GPWv411: UN-Adjusted Population Density (Gridded Population of the World Version 4.11)}&
\RaggedRight{Population density}& 
\RaggedRight{2000; 2005; 2010; 2015; 2020 / 5 years}&
\RaggedRight{0.25 $^{\circ}$}
\\\hline

\RaggedRight{Global Wildfire Information System  (GWIS)}&
\RaggedRight{GlobFire Fire Perimeters (2001-2020)}&
\RaggedRight{Burned Areas}& 
\RaggedRight{2001-2020 / monthly}&
\RaggedRight{500 m}
\\\hline

\RaggedRight{ESA CCI}&
\RaggedRight{MODIS Fire\_cci Burned Area pixel product version 5.1 (FireCCI51)}&
\RaggedRight{Burned Areas ; Fraction of burnable area; Number of patches; Fraction of observed area}& 
\RaggedRight{2001-2020 / monthly averages}&
\RaggedRight{0.25 $^{\circ}$}
\\\hline

\RaggedRight{ORNL DAAC}&
\RaggedRight{Global Fire Emissions Database (GFED.v4) \cite{gfed}}&
\RaggedRight{Burned Areas (large fires only); basis regions (mask)}& 
\RaggedRight{2001-2015 / monthly averages}&
\RaggedRight{0.25 $^{\circ}$}
\\\hline

\end{tabularx}
\caption{Summary of curated datasets, encompassing information on providers, variables, and spatiotemporal resolution.}
\label{tab:table2}

\end{table}

\subsection*{Data acquisition and ARD generation}

\begin{figure}[htbp]
  \centering
  \includegraphics[width=\columnwidth]{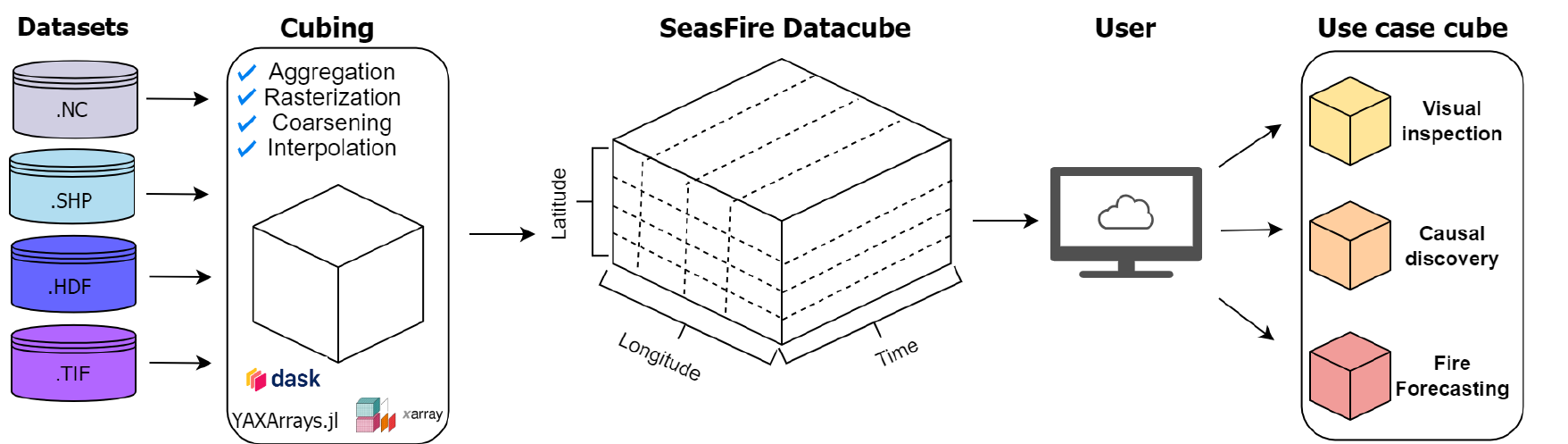}
  \caption{Visualization of the datacube construction workflow. The data are collected from various sources and in different formats. The input data are preprocessed using interpolation, aggregation, coarsening, and rasterization and then appended in the datacube on the corresponding date. After the creation of the datacube, each user can extract task-related machine-learning datacubes.}
  \label{fig:workflow}
\end{figure}

Figure \ref{fig:workflow} illustrates the practical implementation of the aforementioned concept. The flowchart delineates the process, commencing with the collection of relevant datasets. Subsequently, these datasets undergo reprocessing in order to be integrated to as a single datacube.
Users can then load the variables essential for their specific use case, whether on a global scale spanning 21 years or for regional areas and specific time ranges.

We acquired our data from various sources documented in table \ref{tab:table3}, such as the European Centre for Medium-Range Weather Forecasts (ECMWF), Copernicus Climate Data Store (CDS) (\url{https://cds.climate.copernicus.eu/cdsapp\#!/home}). In the same table, we catalog the acquisition methods and formats of each dataset. We collected data covering different temporal resolutions: (a) hourly-daily records for meteorological variables, (b) monthly (e.g. ocean-climate indices), or (c) yearly and sub-yearly (e.g. land cover, and population density) for the study period. 

\begin{table}[!ht]
\def\arraystretch{1.1}%
\centering
\begin{tabularx}{7in}{|p{1in}|p{2.5in}|p{2.1in}|p{0.7in}|}\hline
\textbf{Provider} & \textbf{Dataset} &  \textbf{Acquisition} & \textbf{Format} \\\hline

\RaggedRight{Climate Data Store}&
\RaggedRight{ERA5 hourly data on single levels from 1940 to present \cite{era}}&
\RaggedRight{Web API: \url{https://cds.climate.copernicus.eu/cdsapp\#!/dataset/reanalysis-era5-single-levels?tab=overview}}& 
\RaggedRight{Netcdf}
\\\hline

\RaggedRight{Climate Data Store}&
\RaggedRight{CEMS Global Fire Assimilation System Historical Data \cite{cems_ds}}&
\RaggedRight{Web API: \url{https://cds.climate.copernicus.eu/cdsapp\#!/dataset/cems-fire-historical?tab=overview}}& 
\RaggedRight{Netcdf}
\\\hline

\RaggedRight{Atmosphere Data Store}&
\RaggedRight{CAMS global biomass burning emissions based on fire radiative power (GFAS) \cite{cams_ds}}&
\RaggedRight{Web API: \url{https://ads.atmosphere.copernicus.eu/cdsapp\#!/dataset/cams-global-fire-emissions-gfas?tab=form}}& 
\RaggedRight{Netcdf}
\\\hline

\RaggedRight{NOAA, National Oceanic, and Atmospheric Administration }&
\RaggedRight{Climate Indices: Monthly Atmospheric and Ocean Time Series \cite{oci_soi,oci_censo,oci_ao,oci_nao,oci_ea1,oci_ea2,oci_gmsst,oci_nina34_anom,oci_wp1,oci_wp2,oci_epo,oci_pdo,oci_pna}}&
\RaggedRight{Web Scrapping: \url{https://psl.noaa.gov/data/correlation/}}& 
\RaggedRight{text}
\\\hline

\RaggedRight{ESA CCI}&
\RaggedRight{Land cover classification gridded maps from 1992 to present derived from satellite observations \cite{lc_ds}}&
\RaggedRight{Web Scrapping: \url{https://data.ceda.ac.uk/neodc/esacci/fire/data/burned_area/MODIS/pixel/v5.1/compressed}}& 
\RaggedRight{Netcdf}
\\\hline

\RaggedRight{NASA LP DAAC at the USGS EROS Center}&
\RaggedRight{MOD11C3 v006 \cite{lst_ds} / MCD15A2H v006 \cite{lai_ds} / MOD13C1 \cite{ndvi_ds}}&
\RaggedRight{Web API: \url{https://lpdaac.usgs.gov/product_search/?status=Operational}}& 
\RaggedRight{Netcdf / Shapefile/ Netcdf}
\\\hline

\RaggedRight{RESOLVE Biodiversity and Wildlife Solu\hyp{tions}}&
\RaggedRight{RESOLVE Ecoregions 2017 \cite{resolve_ds}}&
\RaggedRight{Web Scrapping: \url{ https://storage.googleapis.com/teow2016/Ecoregions2017.zip}}& 
\RaggedRight{Shapefile}
\\\hline

\RaggedRight{NASA SEDAC, Socioeconomic Data and Applications Center}&
\RaggedRight{GPWv411: UN-Adjusted Population Density (Gridded Population of the World Version 4.11) \cite{pop_ds}}&
\RaggedRight{Web Scrapping:\url{https://sedac.ciesin.columbia.edu/data/set/gpw-v4-population-density-adjusted-to-2015-unwpp-country-totals-rev11/data-download}}& 
\RaggedRight{Netcdf}
\\\hline

\RaggedRight{Global Wildfire Information System  (GWIS)}&
\RaggedRight{GlobFire Fire Perimeters (2001-2020) \cite{gwis_ds}}&
\RaggedRight{Web Scrapping: 
\url{https://gwis.jrc.ec.europa.eu/apps/country.profile/downloads}}&
\RaggedRight{Shapefile}
\\\hline

\RaggedRight{ESA CCI}&
\RaggedRight{MODIS Fire\_cci Burned Area pixel product version 5.1 (FireCCI51) \cite{fcci_ds}}&
\RaggedRight{Web API: \url{https://cds.climate.copernicus.eu/cdsapp\#!/dataset/satellite-land-cover?tab=form}}& 
\RaggedRight{Netcdf}
\\\hline

\RaggedRight{Oak Ridge National Laboratory (ORNL) Distributed Active Archive Center (DAAC)}&
\RaggedRight{Global Fire Emissions Database (GFEDv4) \cite{gfed}}&
\RaggedRight{Web Scrapping: \url{https://daac.ornl.gov/cgi-bin/dsviewer.pl?ds_id=1293}}& 
\RaggedRight{hdf}
\\\hline

\end{tabularx}
\caption{Dataset acquisition and formats}
\label{tab:table3}
\end{table}

We processed each data variable, by filling in missing values (e.g. population density is provided for every five years, so we forward filled the timestamps of the datacube for the next four years), transforming the data into a consistent format (e.g. shapefiles to netcdf, and missing\_flag as np.nan for all variables), and applying the appropriate land-sea mask to each variable where needed. We then merged all the data, creating a single Zarr file. A detailed data acquisition, aggregation process, and assumptions made are described in detail for each variable in table \ref{tab:table4}. All the acquisition-related information is also compacted and stored as a dictionary on its variable's metadata, so the user can have access directly to the data information.

The main processing techniques are described below and appear on each dataset in table \ref{tab:table4}: 

\paragraph{Aggregation.} Aggregation is the process of creating a more generalized representation of data by either resampling timeseries data or reducing its resolution or granularity. In the case of resampling, we convert hourly or daily data into 8-day averages or cumulative totals, based on the dataset's requirements. The latter method involves grouping data into broader categories, often by applying reduction operations along one or more data dimensions. For example, this might entail reducing the spatial resolution of a land cover dataset from 300 meters per grid cell to approximately 27,000 meters per grid cell.

\paragraph{Interpolation.} Interpolation is a mathematical process used to estimate values for missing or unknown data points by leveraging the available data. It is often employed to fill gaps in a dataset or create a smoother representation of the data. In certain cases, interpolation serves as a technique for resampling or resizing data. We applied nearest-neighbor interpolation, where needed, to ensure compatibility between our datacube's grid and the spatiotemporal datasets we included. 

\paragraph{Filtering.} Filtering is the process of selectively extracting or retaining specific portions of the data while removing unwanted components. It can involve removing noise, outliers, or irrelevant data points based on specific criteria or filters. Filtering was performed in the GWIS dataset, where all the active fire data were removed. 

\paragraph{Rasterization.} Rasterization is the process of converting vector-based data (such as points, lines, or polygons) into a raster or grid format. It involves assigning values or attributes to each cell or pixel in the raster grid based on the characteristics of the original vector data. Rasterization is used in the GWIS dataset for mapping each shapefile to the correct grid cell, as described in the section below.

\subsubsection*{GWIS Burnt areas dataset}\label{subsubsec:1}
The GWIS burnt area data array, derived from the GlobFire dataset \cite{gwis}, has been transformed into a rasterised product. The rasterization process involved several distinct steps, which are outlined as follows:

The GWIS dataset comprises global fire perimeters, distributed in the form of monthly ESRI shapefiles (.shp). Each shapefile contains essential attributes, including a unique fire identification code, initial date (IDate), final date (Fdate), geometric data, and a field indicating whether it corresponds to daily (ActiveArea) or final (FinalArea) burned areas. To quantify the burned area in hectares within this dataset, we leveraged the geopandas library \cite{kelsey_jordahl_2020_3946761}. Subsequently, for the conversion into a datacube format, we employed the xarray library \cite{xarray}. 

\begin{figure}[htbp]
  \centering
  \includegraphics[width=\columnwidth]{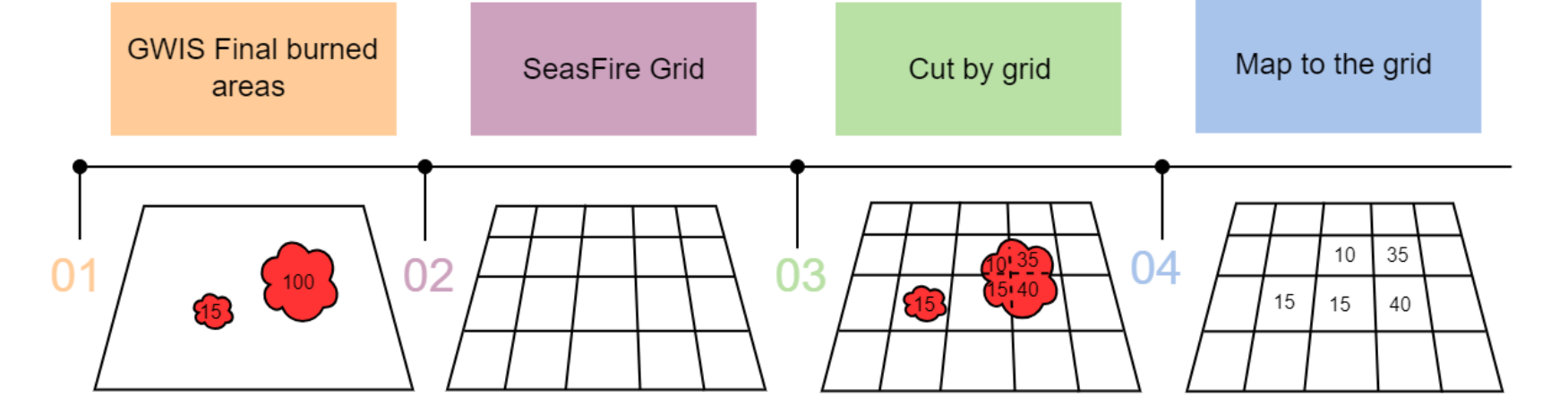}
  \caption{Workflow toy example of rasterization technique. 
}
  \label{fig: rasterization}
\end{figure}

We initially generated yearly raster data to facilitate data processing and the creation of the 8-day granulation and then merged them into a single dataset. During the consolidation of monthly files into yearly datasets, an important observation emerged. Specifically, each monthly file provides information on ignition dates up to approximately the middle of the respective month. For instance, the January shapefile of 2002 includes fires with ignition dates of December 2001, which requires their incorporation into the geodataframe for the year 2001. Given that the dataset spans from 2000 to 2020, this implies that we have data available up to mid-December 2020. Furthermore, we filtered the data to retain only the FinalArea type and disintegrated multipolygons into individual polygons using geopandas. This disintegration step is particularly crucial for the accurate spatial assignment of centroids. The entire process can be delineated through the following steps, as visually represented in figure \ref{fig: rasterization}.

\begin{table} [!ht]
\def\arraystretch{1.1}%
\centering
\begin{tabularx}{7in}{|p{2in}|p{2.5in}|X|X|X|X|}\hline
\textbf{Dataset} & \textbf{Variables} & \textbf{A}& \textbf{I}& \textbf{F}& \textbf{R}\\\hline

\RaggedRight{ERA5 hourly data on single levels from 1940 to present}&
\RaggedRight{Mean sea level pressure; Total precipitation; Relative humidity; Vapor Pressure Deficit; Sea Surface Temperature; Skin temperature; Wind speed at 10 meters; Temperature at 2 meters; Surface net solar radiation; Surface solar radiation downwards; Volumetric soil water levels 1,2,3 and 4}& 
\RaggedRight{\cmark}&
\RaggedRight{\cmark}&
\RaggedRight{\xmark}&
\RaggedRight{\xmark}
\\\hline

\RaggedRight{CEMS Global Fire Assimilation System Historical Data.}&
\RaggedRight{Drought Code; Fire Weather Index}& 
\RaggedRight{\cmark}&
\RaggedRight{\cmark}&
\RaggedRight{\xmark}&
\RaggedRight{\xmark}
\\\hline

\RaggedRight{CAMS global biomass burning emissions based on fire radiative power (GFAS)}&
\RaggedRight{Carbon dioxide emissions from wildfires; Fire radiative power}& 
\RaggedRight{\cmark}&
\RaggedRight{\cmark}&
\RaggedRight{\xmark}&
\RaggedRight{\xmark}
\\\hline

\RaggedRight{Climate Indices: Monthly Atmospheric and Ocean Time Series}&
\RaggedRight{Western Pacific Index; Pacific North American Index; North Atlantic Oscillation; Southern Oscillation Index; Global Mean Land/Ocean Temperature; Pacific Decadal Oscillation; Eastern Asia/Western Russia; East Pacific/North Pacific Oscillation; Nino 3.4 Anomaly; Bivariate ENSO Timeseries; Arctic Oscillation}& 
\RaggedRight{\cmark}&
\RaggedRight{\xmark}&
\RaggedRight{\xmark}&
\RaggedRight{\cmark}
\\\hline

\RaggedRight{Land cover classification gridded maps from 1992 to present derived from satellite observations}&
\RaggedRight{No data; Agriculture; Forest; Grassland; Wetlands; Settlement; Shrubland; Sparse vegetation; bare areas, permanent snow and ice,  Water Bodies}& 
\RaggedRight{\cmark}&
\RaggedRight{\cmark}&
\RaggedRight{\xmark}&
\RaggedRight{\xmark}
\\\hline

\RaggedRight{MOD11C3 v006 / MCD15A2H v006/ MOD13C1}&
\RaggedRight{Land Surface Temperature / Leaf Area Index / Normalized Difference Vegetation Index}&
\RaggedRight{\cmark}&
\RaggedRight{\cmark}&
\RaggedRight{\xmark}&
\RaggedRight{\cmark}
\\\hline

\RaggedRight{RESOLVE Ecoregions 2017}&
\RaggedRight{biomes}& 
\RaggedRight{\xmark}&
\RaggedRight{\xmark}&
\RaggedRight{\xmark}&
\RaggedRight{\cmark}
\\\hline

\RaggedRight{GPWv411: UN-Adjusted Population Density (Gridded Population of the World Version 4.11)}&
\RaggedRight{Population density}& 
\RaggedRight{\cmark}&
\RaggedRight{\cmark}&
\RaggedRight{\xmark}&
\RaggedRight{\xmark}
\\\hline

\RaggedRight{GlobFire Fire Perimeters (2001-2020)}&
\RaggedRight{Burned Areas}& 
\RaggedRight{\cmark}&
\RaggedRight{\xmark}&
\RaggedRight{\cmark}&
\RaggedRight{\cmark}
\\\hline

\RaggedRight{MODIS Fire\_cci Burned Area pixel product version 5.1 (FireCCI51)}&
\RaggedRight{Burned Areas ; Fraction of burnable area; Number of patches; Fraction of observed area}& 
\RaggedRight{\cmark}&
\RaggedRight{\cmark}&
\RaggedRight{\xmark}&
\RaggedRight{\xmark}
\\\hline

\RaggedRight{Global Fire Emissions Database (GFEDv4)}&
\RaggedRight{Burned Areas (large fires only); basis regions (mask)}& 
\RaggedRight{\cmark}&
\RaggedRight{\cmark}&
\RaggedRight{\xmark}&
\RaggedRight{\cmark}
\\\hline

\end{tabularx}

\caption{Variables and processing techniques. A: Aggregation, I: Interpolation, F: Filtering, R: Rasterization}
\label{tab:table4}

\end{table}

\begin{enumerate}
  \item \textbf{Data Processing:} We concatenate monthly files into yearly geodataframes such that each year's file contains all the fire events that started during that period.
  \item \textbf{Create a Grid:} We create a grid according to the spatial dimensions of the datacube [-$90^{\circ},90^{\circ},-180^{\circ},180^{\circ}$] with the step of $0.25^{\circ}$.
  \item \textbf{Split the geometry of fire across grid cells:} Intersect the grid with the geodataframe. For example, figure  \ref{fig: rasterization}, demonstrates how fire events are allocated to each grid cell. In this step we change the coordinate reference system from geographic WGS84 (ESPG:4326) with degree units to an equal area projected WGS84 (ESPG:8857) with units in metres, to calculate the area in hectares. 
  \item \textbf{Assign burned areas to the datacube:} We assign the hectares of each geometry to the grid cell of the SeasFire cube it belongs to, using a rasterio.features.geometry\_mask (\url{https://rasterio.readthedocs.io/en/latest/api/rasterio.features.html}) module.
\end{enumerate}

\section*{Data Records} 

Creating an Earth System Datacube involves a systematic process of acquiring, processing, and organizing diverse Earth observation data from multiple sources and endpoints such as the Climate Data Store ( \url{https://cds.climate.copernicus.eu/cdsapp\#!/home}), Atmosphere Data Store (\url{https://ads.atmosphere.copernicus.eu/cdsapp\#!/home}), National Oceanic, and Atmospheric Administration (\url{https://psl.noaa.gov/data/}), ESA CCI (\url{https://climate.esa.int/en/odp/\#/dashboard}), RESOLVE Biodiversity and Wildlife Solutions (\url{https://storage.googleapis.com/teow2016/Ecoregions2017.zip}, Licensed under  \href{https://creativecommons.org/licenses/by/4.0/}{CC-BY 4.0} ), Socioeconomic Data and Applications Center (\url{https://sedac.ciesin.columbia.edu/data/sets/browse}), Global Wildfire Information System (\url{https://gwis.jrc.ec.europa.eu/apps/country.profile/downloads}), NASA LP DAAC at the USGS EROS Center (\url{https://lpdaac.usgs.gov/product_search/}), Global Fire Emissions Database (GFED) (\url{https://daac.ornl.gov/cgi-bin/dsviewer.pl?ds_id=1293}).  

\begin{figure}[ht]
    \centering
    \includegraphics[width=\columnwidth]{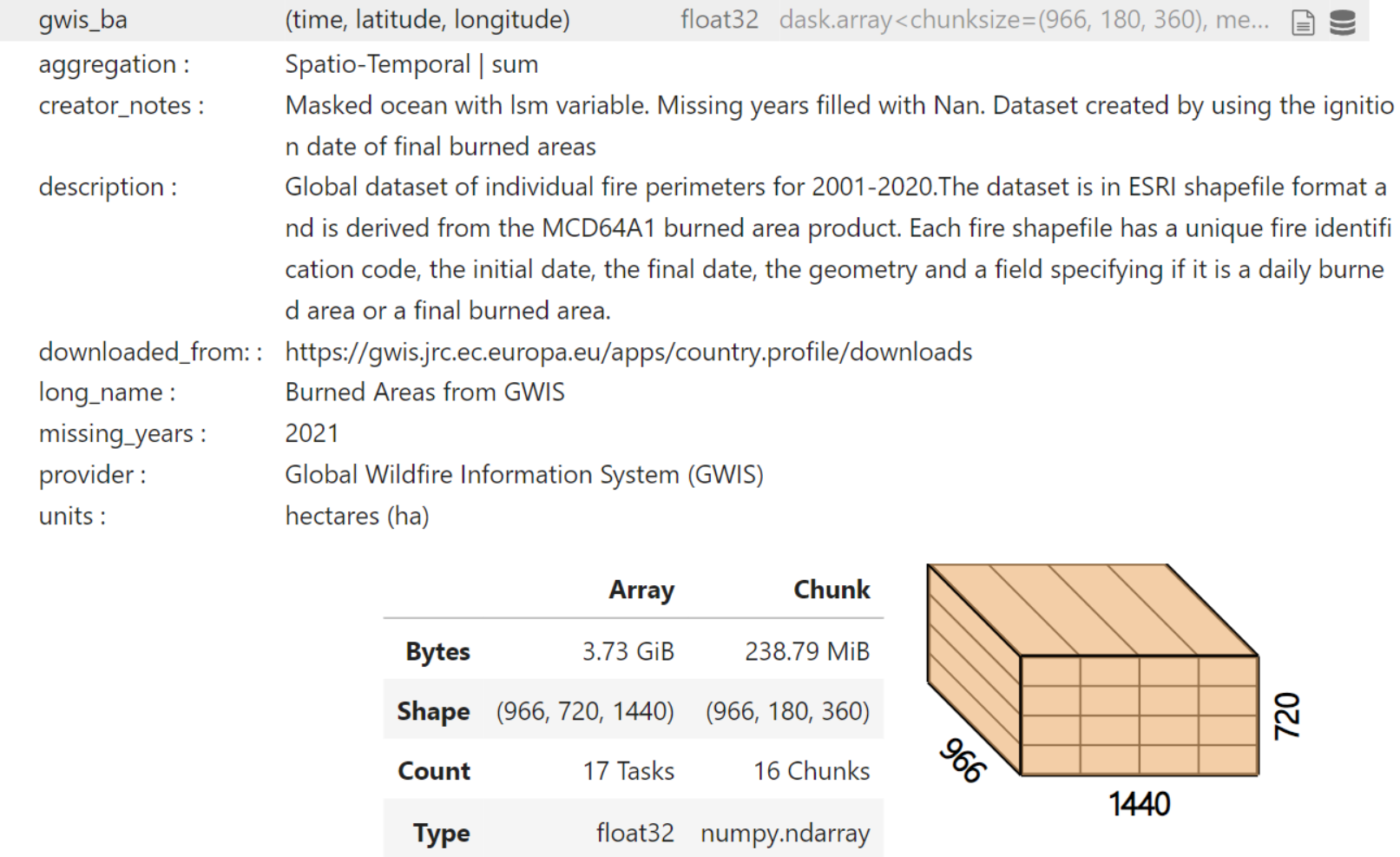}
    \caption{Metadata example of gwis\_ba variable in the datacube. Description of the variable, aggregation performed, units, download link, long name, dataset provider, missing years (if any), and notes for the user.}
    \label{fig:metadata}
\end{figure}

The SeasFire datacube version 3.0 is stored in a .zarr file and can be accessed via Zenodo \href{https://zenodo.org/record/8055879}{https://zenodo.org/record/8055879}.
The dataset's acquisition process is comprehensively described in table \ref{tab:table2}. The original data sources are cited within this paper (refs \cite{oci_soi,oci_censo,oci_ao,oci_nao,oci_ea1,oci_ea2,oci_gmsst,oci_nina34_anom,oci_wp1,oci_wp2,oci_epo,oci_pdo,oci_pna,lst_ds,era,ndvi_ds, lai_ds,gfed,gwis_ds,fcci_ds,resolve_ds,lc_ds}). Every variable includes inherent descriptions, details on aggregation, long names, dataset providers, and user notes, all stored as attributes (metadata) (figure \ref{fig:metadata}) within the Zarr file.

The Zarr format allows optimized chunking for efficient access, storage, and time-based processing of multidimensional gridded data. Each variable is divided into 16 large chunks, structured with dimensions of time (966), latitude (180), and longitude (360), as illustrated in figure \ref{fig:metadata}. Zarr integrates seamlessly with Dask and Xarray, facilitating advanced data analysis while allowing users to work with specific sections of the cube without loading the entire dataset. Extending the datacube with new variables is straightforward, requiring minimal effort if the new data aligns with the existing spatial or temporal axes and compatible format.
The SeasFire datacube can be easily updated with other satellite EO data collections (e.g., MODIS, Sentinel-5P), as well as in the time dimension. As new data will be organized and pre-processed following the protocols presented in this paper, new data streams can be readily included.

\section*{Technical Validation}

The SeasFire datacube gathers and harmonizes validated data records from the respective data providers as shown in table \ref{tab:table5}. To enhance the reliability and accuracy of the SeasFire datacube, a comprehensive technical validation process was conducted, combining analysis and discovery techniques. This validation aimed to assess the quality, consistency, and usability of the datacube for effective wildfire monitoring and analysis. Although wildfire prediction is the primary aim, the data can also be used in the form of individual timeseries. Below we provide some examples to demonstrate the sanity check of some data in the form of data exploration (Visual Inspection), causal links (Causality), and modeling results (Machine Learning). 

Moreover, throughout the development of this datacube, we have committed to continuous refinement and enhancement, resulting in three distinct versions. Each version marks a substantial step forward in terms of quality and utility, and comprehensive changelogs for all are available on Zenodo (\url{https://zenodo.org/record/8055879}) for reference. These versions are a testament to our ongoing commitment to delivering the most valuable and up-to-date resource for the research community.

\begin{table}[ht]
\def\arraystretch{1.1}%
\centering
\begin{tabularx}{7.2in}{|p{2in}|p{1.5in}|p{1.5in}|p{1.5in}|}\hline
\textbf{Dataset} & \textbf{Comparisons with other datasets} & \textbf{Uncertainty estimates} & \textbf{performance check across smaller areas}\\\hline

\RaggedRight{ERA5 hourly data on single levels from 1940 to present}&
\RaggedRight{\cmark}& 
\RaggedRight{\xmark}&
\RaggedRight{\cmark}
\\\hline

\RaggedRight{CEMS Global Fire Assimilation System Historical Data.}&
\RaggedRight{\cmark}& 
\RaggedRight{\xmark}&
\RaggedRight{\cmark}
\\\hline

\RaggedRight{CAMS global biomass burning emissions based on fire radiative power (GFAS)}&
\RaggedRight{\cmark}& 
\RaggedRight{\xmark}&
\RaggedRight{\cmark}
\\\hline

\RaggedRight{Climate Indices: Monthly Atmospheric and Ocean Time Series}&
\RaggedRight{N/A}& 
\RaggedRight{N/A}&
\RaggedRight{N/A}
\\\hline

\RaggedRight{Land cover classification gridded maps from 1992 to present derived from satellite observations}&
\RaggedRight{\cmark}& 
\RaggedRight{\xmark}&
\RaggedRight{\cmark}
\\\hline

\RaggedRight{MOD11C3 v006 / MCD15A2H v006/ MOD13C1}&
\RaggedRight{\cmark}& 
\RaggedRight{\cmark}&
\RaggedRight{\cmark}
\\\hline

\RaggedRight{RESOLVE Ecoregions 2017}&
\RaggedRight{\cmark}& 
\RaggedRight{\xmark}&
\RaggedRight{\cmark}
\\\hline

\RaggedRight{GPWv411: UN-Adjusted Population Density (Gridded Population of the World Version 4.11)}&
\RaggedRight{N/A}& 
\RaggedRight{N/A}&
\RaggedRight{N/A}
\\\hline

\RaggedRight{GlobFire Fire Perimeters (2001-2020)}&
\RaggedRight{\cmark}& 
\RaggedRight{\xmark}&
\RaggedRight{\cmark}
\\\hline

\RaggedRight{MODIS Fire\_cci Burned Area pixel product version 5.1 (FireCCI51)}&
\RaggedRight{\cmark}& 
\RaggedRight{\cmark}&
\RaggedRight{\cmark}
\\\hline

\RaggedRight{Burned Areas from GFED (large fires only)}&
\RaggedRight{\cmark}& 
\RaggedRight{\xmark}&
\RaggedRight{\cmark}
\\\hline

\end{tabularx}
\caption{Overview of dataset's evaluation}
\label{tab:table5}
\end{table}

\subsection*{Visual inspection}

\begin{figure}[!ht]
    \centering
    \includegraphics[width=\columnwidth]{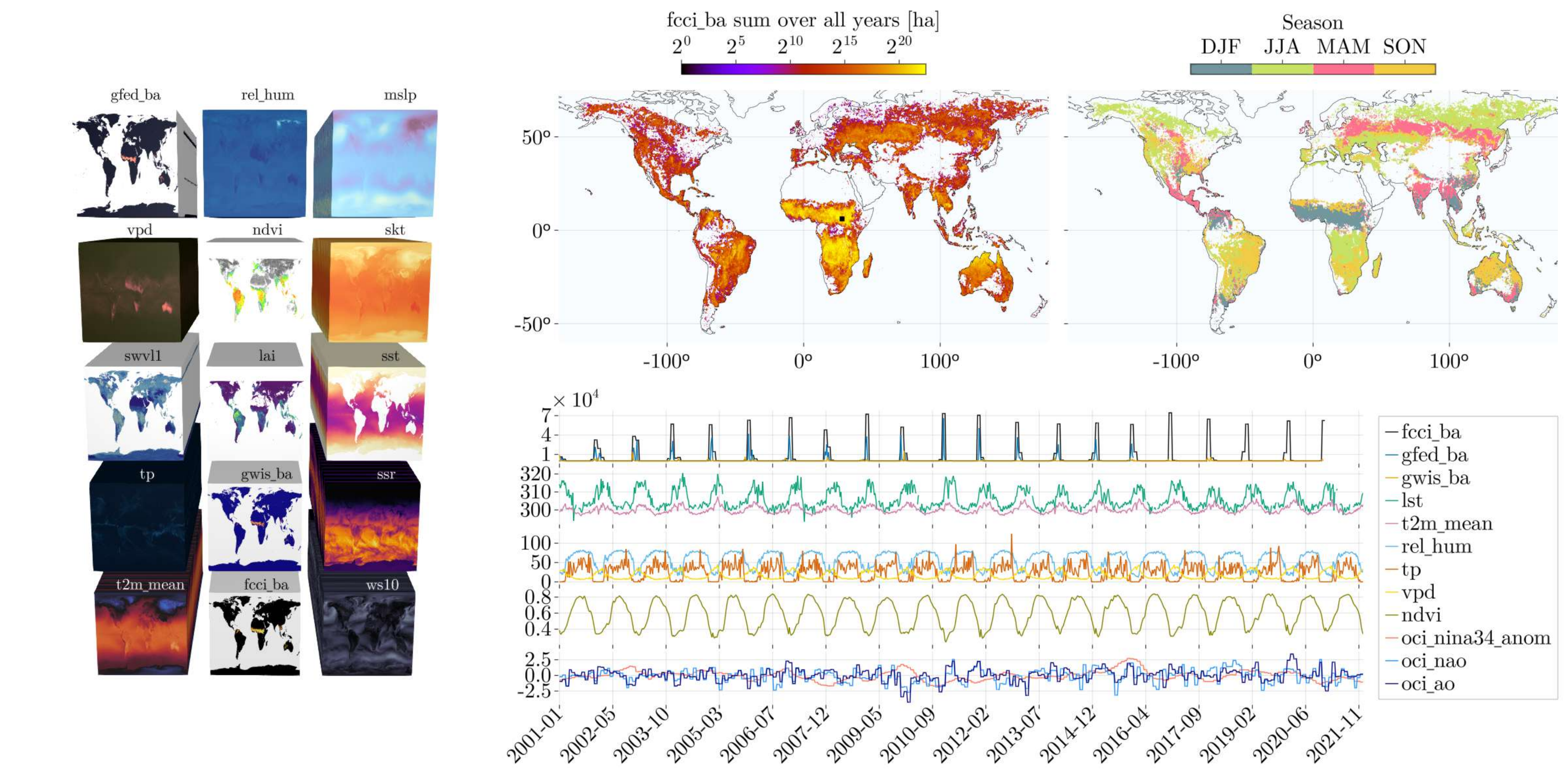}
    \caption{Visual inspection of the datacube that primarily highlights key variables. The left panel displays the variables selected for the subsequent machine learning forecasting approach. The top/middle map illustrates the cumulative burnt area across multiple years, with the black square denoting the location of maximum accumulation. This location serves as the basis for extracting the timeseries displayed at the bottom. Additionally, we also include data on El-Niño (oci\_nina34\_anom), the North Atlantic Oscillation (oci\_nao), and the Arctic Oscillation (oci\_ao), all of which are employed in the subsequent causality analysis. In the top-right map, we present the season with the highest burnt area, represented by the acronyms DJF(Dec-Jan-Feb), JJA(Jun-Jul-Aug), MAM(Mar-Apr-May), and SON(Sep-Oct-Nov).}
    \label{fig:visual_inspection}
\end{figure}

In the realm of data analysis, a preliminary visual inspection plays a pivotal role in unveiling the datacube's underlying characteristics. This process entails a survey of the datacube's variables, with a focus on those used for causality assessment and subsequent machine learning modeling, gaining a deeper understanding of the temporal and spatial patterns of wildfire dynamics around the globe. In figure \ref{fig:visual_inspection}, we included an example of how visual examination reveals trends, seasonal patterns, and relationships within the data, offering valuable clues for the formulation of hypotheses and the design of appropriate modeling strategies. This task here has been done using the plotting library Makie.jl\cite{DanischKrumbiegel2021}. 

\subsection*{Causality}

Causal analysis can be a method for validating the quality of datasets by confirming theoretical Earth system science cause-and-effect relationships among variables, strengthening the overall reliability of the datacube. To demonstrate this, we conducted an experiment linking climate, meteorology, and burned areas in the European Mediterranean and Boreal region over two decades, using the SeasFire datacube. The PCMCI method \cite{Runge2019} for causal discovery, designed for timeseries data, employs the Peter-Clark (PC) algorithm that identifies all the causal graphs that are consistent with the available data, and Momentary Conditional Independence (MCI) test to further assess causal relationships. The MCI test takes into account factors like auto-correlation and erroneous edge detections, enhancing the accuracy of causal discovery. Tigramite (\url{https://tocsy.pik-potsdam.de/tigramite.php}), a versatile Python framework, supports PCMCI and other methods for causal discovery, even accommodating nonlinearities. However, geoscientific timeseries, like those involving oceanic and atmospheric processes, pose challenges due to non-Gaussian noise \cite{Runge2019_1}, making it difficult for statistical tests to capture complex nonlinear relationships between variables.

\begin{figure}[ht]
    \centering
    \includegraphics[width=\columnwidth]{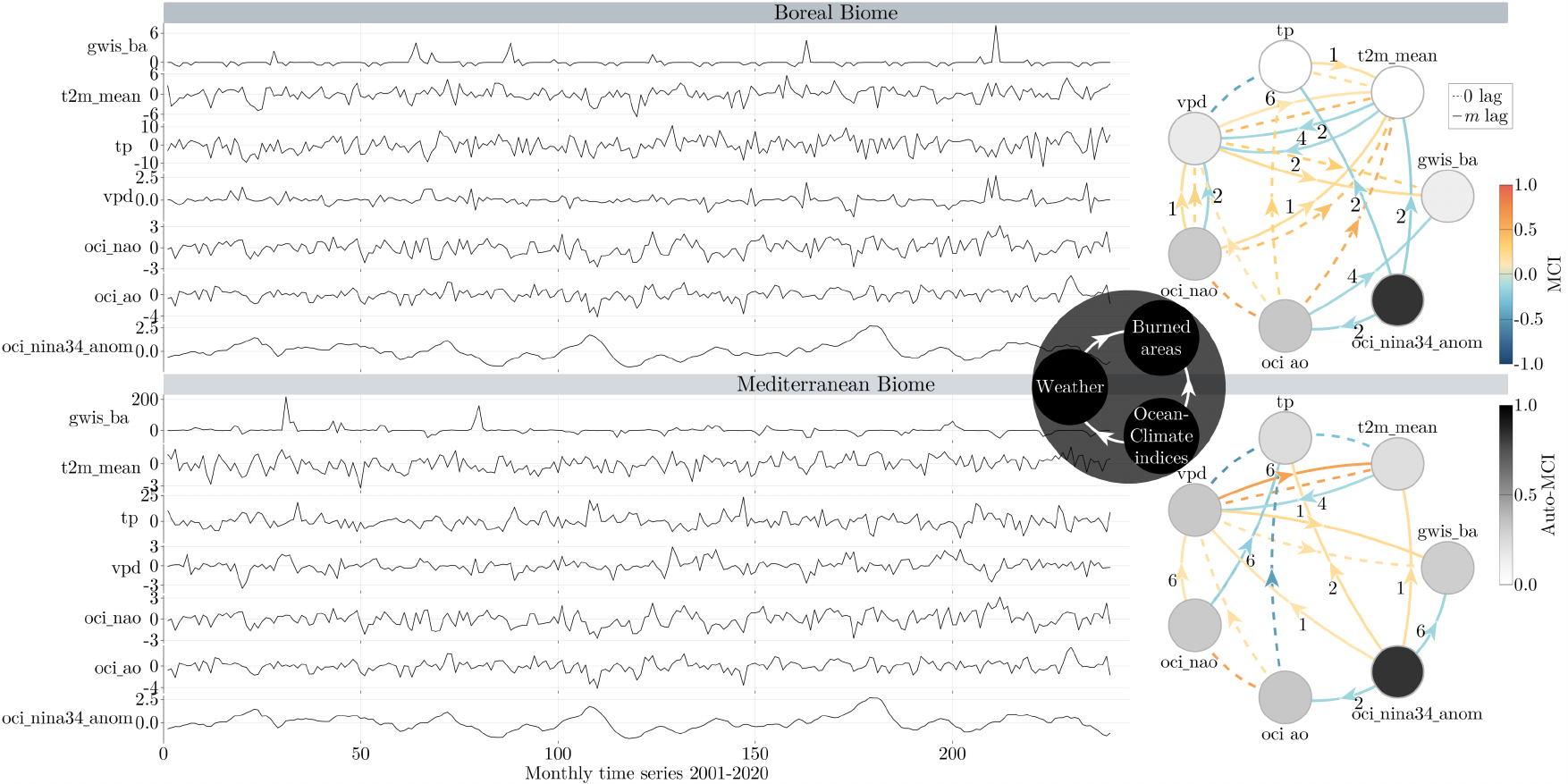}
    \caption{Timeseries, causal order and PCMCI causal discovery graphs for the Mediterranean and Boreal forests. The color of each node indicates its self-correlation (the relationship with itself over time). In the causal networks, each node on the right side corresponds to a variable, as seen in the timeseries on the left. The color of the links indicates the partial correlation value between variables, which reveals the direction and strength of the inferred causal connection between them. For lagged connections, you can find the time delay (in months) indicated by small labels on the curved arrows, while dashed lines denote instantaneous causal connections occurring without any time delay. The illustrated key climatic and environmental variables include: oci\_ao (Arctic oscillation), oci\_nao (North Atlantic oscillation), oci\_nina34\_anom (El-Niño at 3.4 region), vpd (vapor pressure deficit), temp\_mean (Mean air temperature at 2 meters), tp (total precipitation), and gwis\_ba (GWIS burned areas). }
    \label{fig:causal_graphs}
\end{figure}

Our case study employed linear partial correlation tests (ParCorr) with PCMCI under specified assumptions \cite{Pearl2009, Spirtes2001}, via preprocessing, parameter tuning, and causal network learning. The causal graphs are presented in figure \ref{fig:causal_graphs}, and illustrate the causal relationship (positive or negative), strength (0 to |1|), and time lag (0 to N months) of that relationship. The initial preprocessing involved transforming variables including Burned Areas, North Atlantic Oscillation, Arctic Oscillation, El-Niño in 3.4 region, precipitation, vapor pressure deficit, and temperature into monthly timeseries while ensuring their stationarity through anomaly calculation for all variables except oscillations and El-Niño in 3.4 region. For parameter tuning, a maximum lag of 6 months was chosen to account for seasonal changes, and an alpha significance threshold of 0.05 was applied to the independence test. To improve the outcome and minimize the presence of spurious links, we established a causal order as it appears in the grey circle in the middle of figure \ref{fig:causal_graphs}. In this order, the possible causal links are the following. The ocean climate indices were considered causal for themselves, meteorology, and burned areas. Meteorology was deemed causal for itself and for burned areas only, and lastly, the burned areas are not considered causal for any of the variables.

The graphical analysis underscores the preeminent role of vapor pressure deficit timeseries, indicative of dryness \cite{Li2023}, in influencing burned areas for both the Euro-Mediterranean and Euro-Boreal biomes. Notably, lags of 0 and 1-month exhibit significance for the Euro-Mediterranean, while lags of 0 and 2-months are influential for the Euro-Boreal biome. A closer look at the graphs reveals the following insights. The El-Niño (node Nino\_3.4\_anom) timeseries demonstrates a substantial self-memory (high Auto-MCI) of its past patterns, although this doesn't necessarily imply the predictability of the timeseries itself. As anticipated, a negative association between vapor pressure deficit (indicating dryness) and precipitation exists as contemporaneous links (lines without arrows), along with a positive correlation between vapor pressure deficit and temperature at 0 and 6-month lag time. Notably, the Arctic oscillation (oci\_ao) and the North Atlantic oscillation (oci\_nao) are strongly positively correlated in this context at 0-month lag. \cite{almendra-martin_influence_2022,climategov}. Lastly, concerning the lagged nonlinear causalities (lines with arrows-ParCorr test), across the Euro-Mediterranean region, Benassi et al. \cite{Benassi2021} confirm the outcomes of our analysis, wherein a positive El-Niño influence augments rainfall, while the same influence in boreal regions leads to negative temperature anomalies. As expected \cite{climategov}, Arctic Oscillation exhibits a leading role in negatively affecting precipitation at 0-month lag within Mediterranean forests, whereas it positively influences temperature contemporaneously in Boreal regions. These causal findings, in alignment with the existing literature, serve as compelling evidence that the timeseries data under examination are indeed reliable.

\subsection*{Machine Learning Modeling}

In this section, we showcase the potential of the SeasFire cube for Machine Learning applications. Having the cube that contains historical data related to the wildfire drivers and wildfire effects (burned areas, emissions), we can model the connections between the root causes and their resulting effects. 

In the process of translating data from a datacube into a machine learning task, several critical steps must be carefully navigated, which are elaborated upon as follows. Initially, one must clearly delineate the task's objectives and desired outcomes, while simultaneously devising a strategy for data sampling, mindful of the temporal and spatial aspects. A robust evaluation split is pivotal, as it determines how the dataset will be divided into training, validation, and test subsets, steering clear of potential data leakage. Setting an appropriate baseline, often in the form of a simple model, provides a reference point for advanced model comparisons. Lastly, defining relevant evaluation metrics is crucial, ensuring that the chosen measures accurately reflect the task's real-world implications, thus paving the way for effective model development and evaluation. We demonstrate these five steps in the following section where we define burned area pattern forecasting as a segmentation task. 

\subsubsection*{Burned Area Pattern Forecasting as a Segmentation task}

\begin{figure}[ht]
    \centering
    \includegraphics[width=\columnwidth]{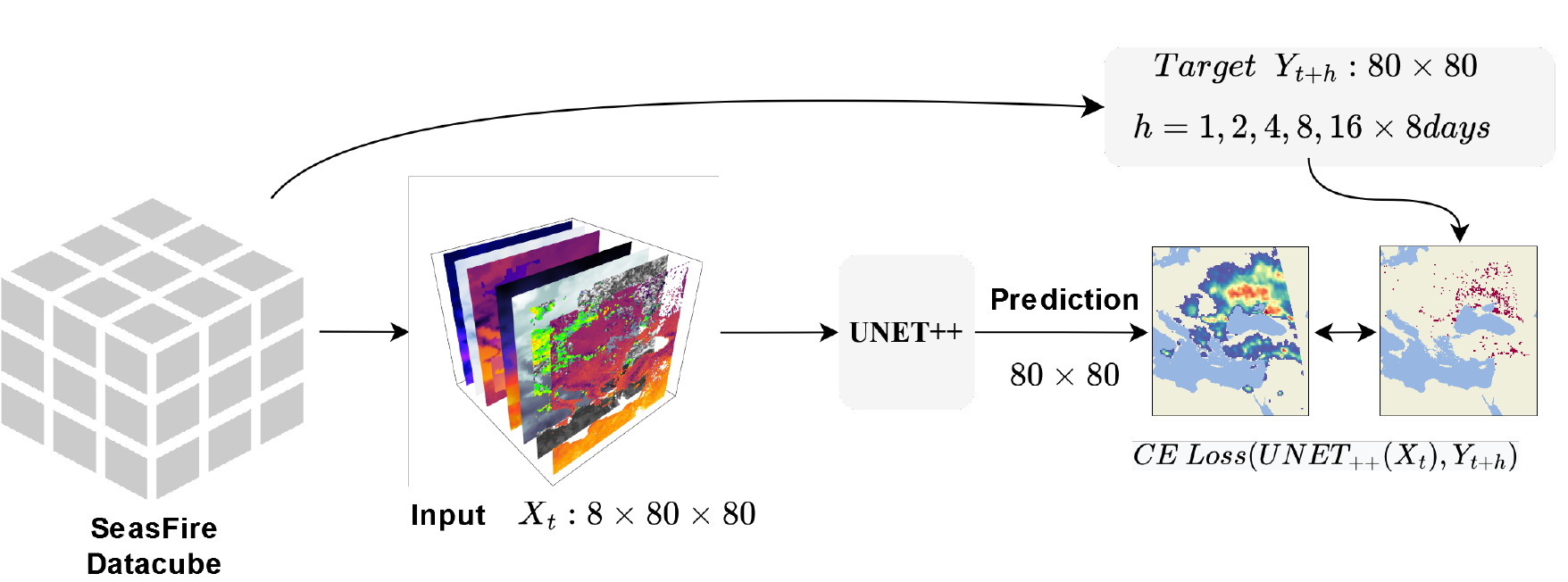}
    \caption{Performance of the U-Net++ model for predicting at different forecasting horizons. In red text, the performance of the mean seasonal cycle baseline.}
    \label{fig:ml_pipeline}
\end{figure}

\paragraph{Formulation and setup.} Burned area pattern forecasting can be defined as a segmentation task and in fact, this has been demonstrated by existing work on the SeasFire cube \cite{Prapas_2022, Prapas_2023}. 
Following \cite{Prapas_2022}, 8 variables from the cube are used as input features and the target is derived from the GWIS burned area variable. Input, target pairs have a spatial dimension of 80 x 80 pixels or 20 $^{\circ}$ x 20 $^{\circ}$. 
The 8 input variables are stacked in an 8 x 80 x 80 tensor and fed as input to a U-Net++ model \cite{zhou_unet_2018}.  
As such, the U-Net model is trained to predict the presence of burned areas at time $t+h$ using the input variables at time $t$. For each pixel, the burned area mask signifies the presence or absence of a burned area. 
This simplifies the machine learning task, by not taking into account the actual burned area size of the cell. The pipeline is shown in figure \ref{fig:ml_pipeline}. The forecasting lead time $h$ is in multiples of 8 days, which is the temporal resolution of the dataset. 
A different model is trained for each of the four different forecast lead times ($h$ = 8, 16, 32, 64 days).

To account for the imbalance between the negative (not burned) and the positive (burned) class, we train on patches that include at least one burned area pixel. This discards about half of the examples, mainly on polar regions or in the sea, ending up with 44,255 out of the 134,622 patches. 
Dealing with a forecasting task, we do a time split using years from 2002 to 2017 for training (39,265 examples), 2018 for validation (2,426 examples), and 2019 for testing (2,564 examples). Among the total pixels of those datasets, only about 1.6\% are of the positive class (burned). As an evaluation metric, we use the Area Under the Precision-Recall Curve (AUPRC), which is appropriate as a measure of skill for an imbalanced target class. As a statistical baseline, we use the weekly mean seasonal cycle. We calculate the temporal average of burned area sizes for each one of the weeks in a year, on the training years (2002 - 2018). Each patch on the validation and test set is compared to an equal-dimension patch that contains the average of the respective week of the historical dataset. The AUPRC of the weekly averages are also reported. 
 
We use the EfficientNet-b1 encoder \cite{tan2019efficientnet} as a backbone of the U-Net++ model and train for 30 epochs with the cross-entropy loss, which is commonly used for segmentation tasks. 

\paragraph{Results and discussion.} 

\begin{figure}[ht!]
    \centering
    \includegraphics[width=0.5\columnwidth]{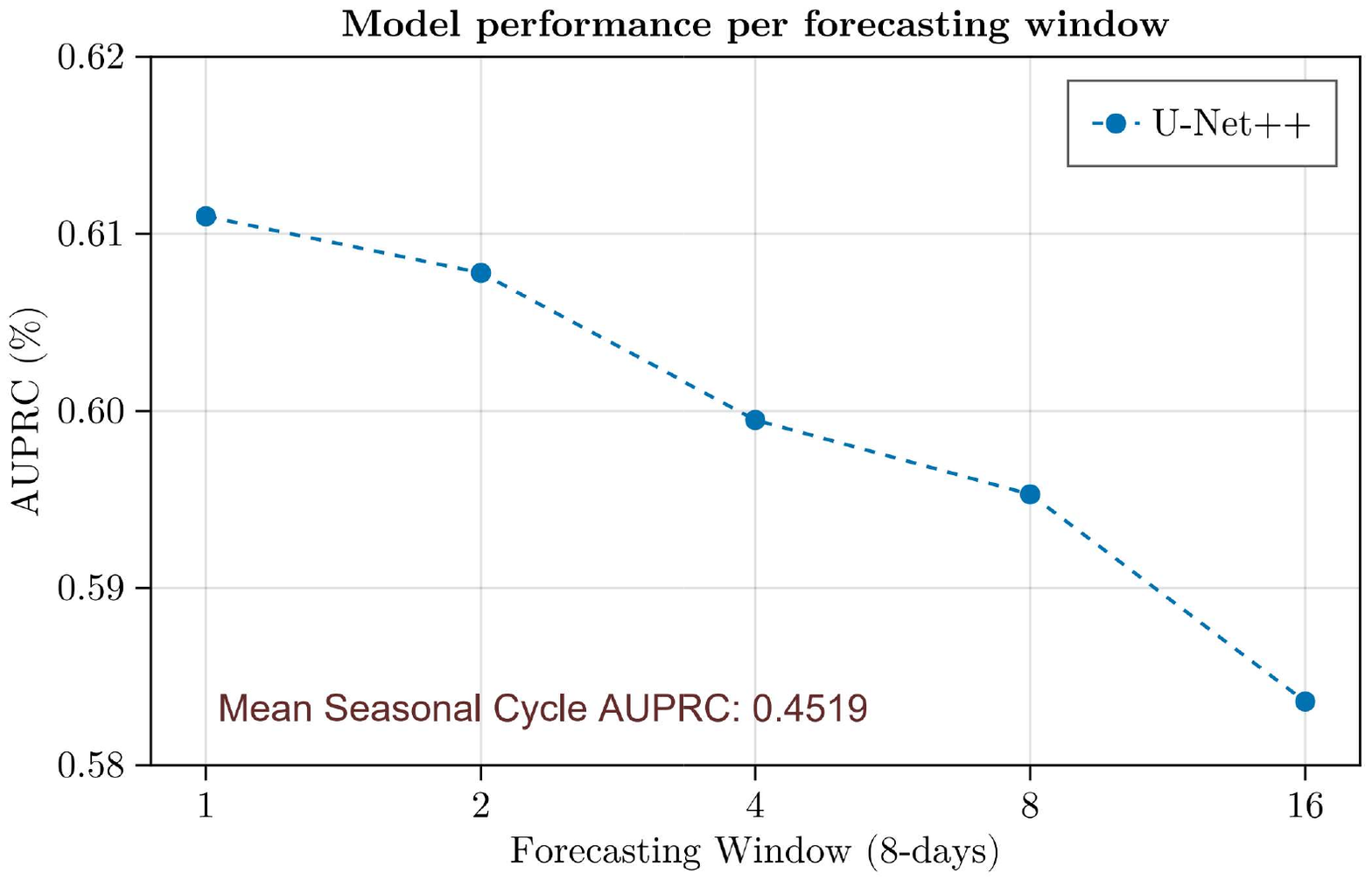}
    \caption{Performance of the U-Net++ model for predicting at different forecasting horizons. In red text, the performance of the mean seasonal cycle baseline.}
    \label{fig:unet_results}
\end{figure}

\begin{figure}[ht!]
    \centering
    \includegraphics[width=\columnwidth]{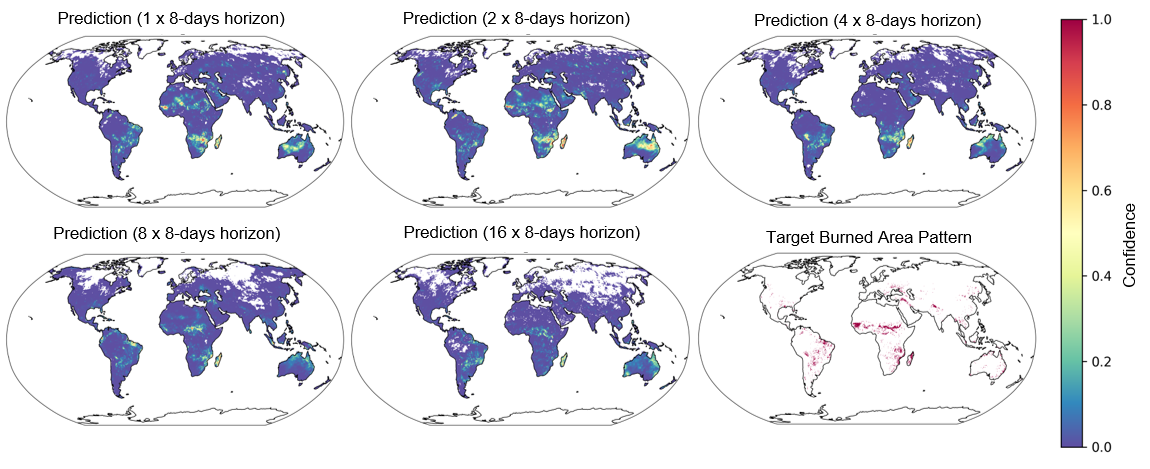}
    \caption{Comparison between the prediction of the model and the target burned area pattern for the different forecasting horizons. The target for all the predictions comes from the datetime 2019-11-01 from the datacube, while the input for the different predictions is shifted backward by several 8-days indicated by the forecasting horizon.}
    \label{fig:burned_area_forecasts}
\end{figure}

Figure \ref{fig:unet_results} shows that the performance in terms of AUPRC drops as the forecasting horizon increases. However, even for predicting about 4 months in advance (16 x 8-days), the performance remains much higher than the mean seasonal cycle baseline, demonstrating an AUPRC of 0.4519 on the test set. Qualitatively, figure \ref{fig:burned_area_forecasts} shows that the model predictions match very well the pattern of the target, with lower confidence when predicting at larger forecasting horizons. Large-scale patterns of the target burned area are well captured (e.g. patterns in the African Savannahs, and the American continent). Further investigation is needed to evaluate the prediction skill at regional levels. We can also notice some wrong predictions like in the south of the Middle East, where the burned area is mistakenly predicted up to about a month in advance.

In short, this work demonstrates that DL models can skilfully predict global burned area patterns from a snapshot of fire driver input data valid from 8 days to more than 4 months in advance. This motivates the use of DL for burned area forecasting and calls for future work that is facilitated by the SeasFire cube.

\subsubsection*{Other Potential Machine Learning Tracks}

Through this SeasFire cube demonstration, we aim to ignite the curiosity of the research community in the challenging and highly significant realm of global wildfire modeling, which stands as a complex and pressing task of our era. The datacube has a unique set of variables that allow us to explore and study wildfire dynamics in many different ways. For instance, the inclusion of time-series data related to oceanic indices empowers us to analyze spatiotemporal teleconnections, unraveling the intricate relationships between oceanic phenomena and wildfires. Furthermore, by incorporating human-related factors like population density and land cover, we can gain valuable insights into how human activities impact wildfires across time and space. Additionally, the datacube encompasses crucial information on carbon emissions resulting from wildfires, providing a valuable resource for studying the profound implications of wildfires on the global carbon cycle.

Machine learning holds immense potential across various fronts in advancing our understanding of wildfires. Firstly, by harnessing machine learning, we could improve the prediction of critical variables such as vegetation growth, drought conditions, and fire weather patterns on sub-seasonal to seasonal timescales, developing early warning systems and enabling more proactive wildfire mitigation efforts. Secondly, shifting from traditional segmentation to regression modeling provides an alternative avenue for comprehending the continuous relationships between variables, resulting in more precise predictions of wildfire occurrences. Moreover, machine learning can extend its reach to forecast wildfire emissions, offering a comprehensive assessment of their environmental impact, a vital aspect for evaluating their effects on air quality and the global carbon cycle. Additionally, by incorporating teleconnection indices and causality analysis, we gain deeper insights into the complex web of interactions between climatic factors and wildfires, enhancing our predictive and explanatory capabilities. Lastly, enhancing explainability in these models is crucial, as it can help researchers and policymakers better interpret and trust the insights derived from these models. explainable AI techniques can make the modeling process more transparent and actionable.

However, it's essential to acknowledge certain limitations. The availability of data is a primary constraint. Climate change operates on larger temporal scales, and having only 21 years of data may limit our ability to judge long-term performance and generalize findings to future climate scenarios. The spatial resolution of 0.25 $^{\circ}$ enables the capture of expansive large-scale patterns and global trends, offering a comprehensive view of macroscopic environmental dynamics. However, this resolution may pose limitations in accurately representing localized variations. The temporal resolution, characterized by an 8-day interval, might be insufficient for modeling scenarios that require precise tracking of daily variations, as is often essential in the case of fire spread dynamics. Additionally, the SeasFire datacube, while a valuable resource, may not encompass all the variables necessary to capture the full complexity of wildfire dynamics. These limitations emphasize the need for ongoing data collection efforts and a comprehensive understanding of the intricacies of global wildfire modeling.

\section*{Usage Notes}

The Seasfire datacube is available for unlimited use under the Creative Commons License 4.0 International. We strongly recommend that users access the cube using multidimensional arrays, for instance, through Xarrays (Python) or YAXArrays.jl (Julia). Additionally, any customized Python or Julia code employed in the analyses of the SeasFire datacube is accessible in the accompanying GitHub repository. Users are encouraged to utilize this code as a foundational resource for their individual analyses. Notebook examples are available for both programming languages (\url{https://github.com/SeasFire/seasfire-datacube/tree/main}). The datacube can be accessed both locally and in a cloud environment. The choice between these two options should be contingent upon the user's available computational resources. It is essential for users to verify that their local or cloud environment possesses adequate Random Access Memory (RAM) capacity to accommodate the processing requirements of the datacube.
The specific RAM requirements may vary depending on the nature of the analyses performed, but as a general guideline, approximately 4 GB of RAM is needed for loading a single variable. The data within SeasFire has undergone preprocessing to a certain extent, which is well documented in the medatada of each variable. Nevertheless, users may find it necessary to conduct further data cleaning, normalization, or transformation procedures depending on their specific research questions.

The methodology employed in the development of the datacube involves the collection and processing of a diverse array of atmospheric, climatological, and socioeconomic variables. It is advisable for users to familiarize themselves with these methodologies to gain a comprehensive understanding of the datacube.
The SeasFire datacube presents the opportunity for potential integration with other relevant datasets, and we encourage users to consider the possibility of incorporating additional data sources to enhance their analyses. Researchers should exercise vigilance regarding potential pitfalls, such as the presence of missing or inconsistent data, which are explicitly documented in the metadata associated with each variable.

\section*{Code availability}

All code necessary for the technical validation will be publicly available on GitHub (\url{https://github.com/SeasFire/seasfire-datacube-paper}). Code regarding working with the SeasFire datacube in both Python and Julia is also provided in the GitHub repository (\url{https://github.com/SeasFire}).

\bibliography{sample.bib}

\section*{Acknowledgements} 

The creation of this datacube has been funded by the European Space Agency (ESA), in the context of ESA Future EO-1 Science for Society Call. Project supported by ESA's Network of Resources Initiative.

\section*{Author contributions statement}

Conceptualization: I.Prapas, N.C., I.Papoutsis; Data curation: I.K., L.A.; Implementation: I.K., L.A., I.Prapas; Technical Validation: I.K., L.A., I.Prapas; Data visualization: I.K., L.A.; Writing - Original Draft: I.K., A.A.; Writing - Review, and Editing: All authors; Supervision: I.Prapas, N.C., I.Papoutsis; Funding acquisition: N.C., I.Papoutsis

\section*{Competing interests}

The authors declare no competing interests.

\end{document}